# An evolutionary model of animats with VLM-based subjective evaluation

Shota Miyazaki[1], Takaya Arita[1] and Reiji Suzuki[1]

[1]Graduate School of Informatics, Nagoya University, Japan
(Tel : +81-52-789-4716; E-mail: miyazaki.shouta.a4@s.mail.nagoya-u.ac.jp)

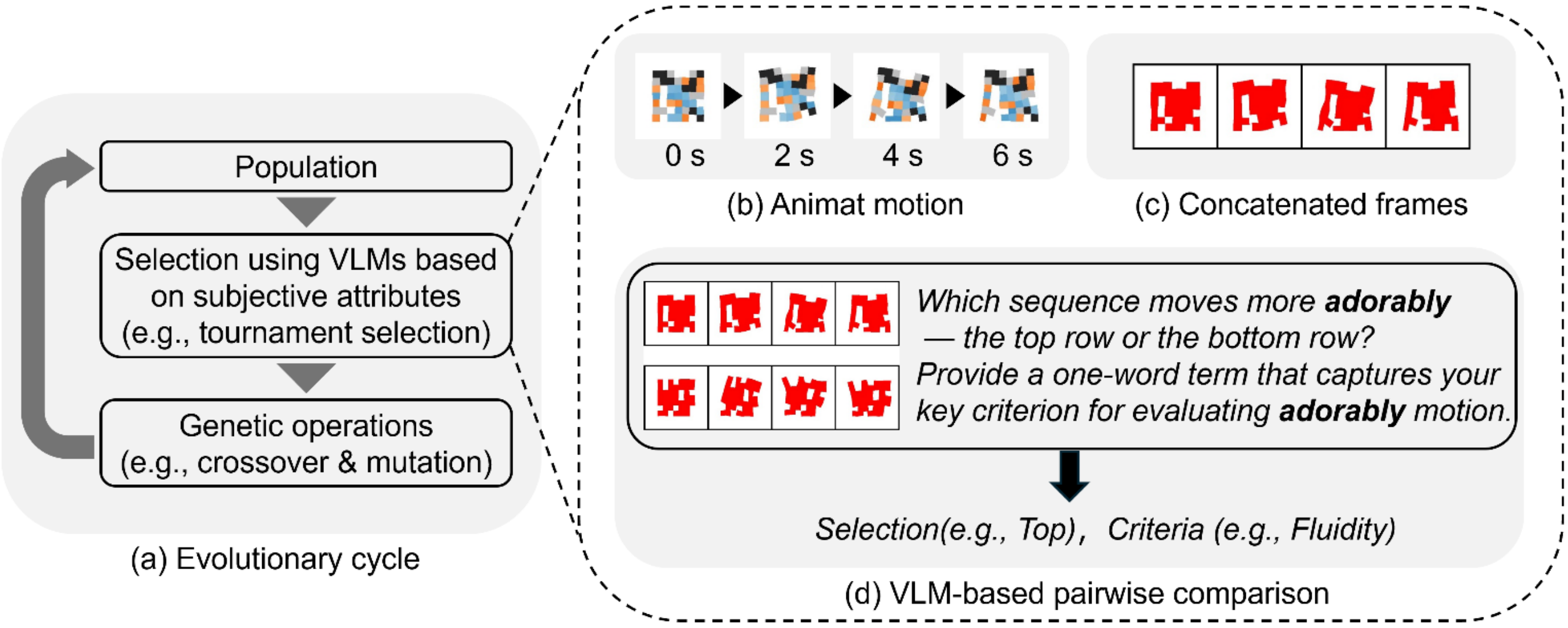

**Fig. 1** Overview of the proposed model. (a) Evolutionary cycle. (b) Example of animat motion. (c) Example of a sequence image. (d) Pairwise comparison based on subjective evaluation by the VLM

## ABSTRACT

In this study, we propose a framework that incorporates subjective evaluations provided by a Vision–Language Model (VLM) into the fitness evaluation and selection processes of a genetic algorithm. As the target of evolution, we employ virtual soft robots with flexible morphologies and locomotion and present the VLM with sequence images representing the locomotion of two individuals. Selection is performed via pairwise comparisons based on subjective evaluation terms such as adorably and weirdly. The outcomes of these comparisons are used as selection pressure within the genetic algorithm, enabling the simultaneous evolution of morphology and locomotion. Experimental results demonstrate that subjective selection by the VLM accelerates population convergence compared to random selection, while also giving rise to distinctive morphologies and motions corresponding to each evaluation term. An auxiliary experiment with human participants further showed that, although individual pairwise choices only partly agreed with the VLM selections, the resulting morphological and locomotion tendencies were qualitatively similar and repeated human evaluations imposed noticeable fatigue. Moreover, the observation that similar evolutionary outcomes emerged across different evaluation terms suggests that the VLM does not apply these terms in a purely literal manner but instead decomposes them into multiple internal evaluation criteria when making judgments. This work visualizes the evolutionary process through which subjective linguistic expressions are mapped onto embodied phenotypes and provides a foundational framework for analyzing the structure of subjective judgment in VLMs. The proposed approach is expected to contribute to new developments in evolutionary computation and artificial life research based on subjective evaluation.

---

This manuscript has been accepted for publication in *Artificial Life and Robotics* following peer review. This preprint is not the Version of Record. The DOI and URL link to the published version of the article on the journal website are provided below.

---

## 1 INTRODUCTION

Many engineering and social problems can be formulated as the minimization or maximization of an evaluation (objective) function, and a wide variety of optimization methods have therefore been proposed to date. However, there also exist problems for which such a formulation is difficult. For example, optimization based on evaluations that depend on human subjectivity or emotions exhibits substantial individual differences and strong context dependence, making it challenging to define an explicit evaluation function and perform optimization.

As a framework that avoids explicitly designing an evaluation function based on such subjective criteria, Interactive Evolutionary Computation (IEC) has been proposed [1]. IEC incorporates human input into the selection of parent individuals in evolutionary computation and advances the search process through selections based on users' subjective preferences. Dawkins' Biomorphs demonstrated that subjective selection pressure arising from observers' preferences can generate morphological diversity, thereby illustrating the potential of exploration driven by subjective evaluation [2].

However, because IEC requires evaluating many individuals over many generations, evaluator fatigue constitutes a major bottleneck. To address this fatigue problem, approaches such as reducing the number of evaluations [3] and improving the efficiency and scalability of evaluation input through web-based systems involving multiple participants [4] have been investigated. Nevertheless, even with these improvements, they have not led to a fundamental solution to the core problem that evaluations performed by humans incur a high cost.

In recent years, with the advancement of Large Language Models (LLMs), LLM-as-a-judge has been proposed as an approach to replace human evaluators. This approach utilizes LLMs to automatically assess the quality of generated text or model outputs based on given criteria [5]. LLM-as-a-judge is significant in that it can automate comparative evaluations based on context-dependent criteria in situations where human evaluation is costly and difficult to scale. Indeed, prior studies have reported that powerful LLMs can exhibit a high degree of agreement with human preferences, as well as the existence of biases in such evaluations and analyses thereof [5]. Moreover, the intersection between LLMs and evolutionary computation has continued to expand. Methods that introduce LLMs as mutation operators have been proposed [6], as well as evolutionary approaches for optimizing prompts using LLMs [7].

Furthermore, the use of Vision-Language Models (VLMs) makes it possible to automate comparative evaluations even for targets that involve visual information such as shape and motion. To investigate how VLMs understand the linkage between vision and language within their cognitive capabilities, tasks in which VLMs are asked to select images have been employed. Results have shown that VLMs exhibit tendencies corresponding to sound symbolism (the so-called "bouba/kiki" effect [8]) observed in humans [9], and that they share not only sound-symbolic associations but also aspects of color perception with humans [10]. Furthermore, methods using foundation models such as CLIP have been proposed to search for artificial life models whose visual behaviors match linguistically specified descriptions [11]. Consequently, the development of optimization methods based on VLMs and a deeper understanding of their underlying mechanisms have become increasingly important.

We aim to propose a framework that incorporates subjective evaluation by a VLM into evolutionary computation, enabling the evolution of animats whose morphologies and behaviors reflect given subjective linguistic expressions. By analyzing the resulting evolutionary processes and phenotypes, we further aim to elucidate how VLMs internally map such expressions onto embodied representations.

To efficiently and reliably evaluate evolutionary targets that are computationally expensive and difficult to compare directly, such as motion represented in video, we designed a method in which the VLM performs step-by-step reasoning and evaluation based on sequences of snapshots arranged for pairs of individuals.

We implemented animats (virtual soft robots) with flexible morphologies capable of expressing subtle motion patterns using Evolution Gym. We conducted evolutionary experiments in which populations of these animats evolved under selection pressure derived from subtle, subjective linguistic expressions such as *adorably* and *weirdly*. In addition, to provide an initial comparison with human subjective perception, we conducted an auxiliary human-subjective IEC experiment for selected abstract evaluation terms and examined the overlap and differences between VLM-driven and human-driven evolutionary outcomes. By analyzing the evolutionary dynamics under these conditions, we discuss how differences in subjective linguistic expressions are reflected in the emergent morphologies and locomotion patterns of animats.

## 2 METHOD

### 2.1 Overview

An overview of the proposed method is illustrated in Figure 1. As the target of evolution, we employ Evolution Gym, a virtual soft robot benchmark, and evolve both the morphology and locomotion of animats. The reason for using animats as evolutionary targets is that abstract information such as subjectivity can be manifested not only in static appearance but also in temporal characteristics, including deformation process over time. This enables us to address scenarios in which subjective attributes emerge through dynamic locomotion rather than solely through external form.

Whether an individual is considered adaptive is determined based on subjective evaluations provided by a VLM, such as assessments of adorable or weird. Specifically, the locomotion of animats is evaluated via pairwise comparisons based on snapshot images of multiple individuals presented simultaneously to the VLM (Fig. 1d). Pairwise comparison is adopted because it is difficult to assign absolute subjective scores to highly abstract targets such as the motion of animats; by presenting multiple candidates simultaneously and asking the model to compare them, more nuanced and consistent evaluations can be obtained. Static images are used instead of videos to reduce the computational cost associated with repeated inference, which is a critical issue in applications to evolutionary computation. By inputting a single image that aggregates snapshot views of multiple individuals, dynamic characteristics such as motion can be evaluated from a single image. In addition, previous work has reported that, for understanding motion, sequences of discrete snapshots can yield better performance than continuous video input [12]. The evolution of animats is implemented by introducing VLM-based subjective evaluations into the fitness evaluation and selection process of a genetic algorithm (Fig. 1a).

### 2.2 Animats

The animats are implemented using the virtual soft-robot benchmark Evolution Gym [13]. Each animat has a body composed of square blocks, referred to as voxels, arranged within a $W \times W$ grid (Fig. 2a). There are four types of voxels: two static blocks, rigid and soft, and two dynamic blocks, the horizontal actuator and vertical actuator (Fig. 2b). Animats move by expanding and contracting the actuator voxels. This deformation is realized by allowing each voxel to change the length of its edges at every time step. The edge length $L(t)$ is computed using the periodic function defined below.

$$L(t) = \frac{\sin\left(\frac{\pi t}{30} + \varphi\right) + 1}{2} + 0.6, \tag{1}$$

where, $t$ denotes the time step in the virtual environment, and $\varphi$ represents a phase parameter taking values in the range $0 \leq \varphi \leq 2\pi$. The genotype of an animat is defined as a one-dimensional list consisting of integers corresponding to voxel types (empty: 0, rigid: 1, soft: 2, horizontal actuator: 3, vertical actuator: 4) and, for voxels that are actuators, real-valued parameters corresponding to the phase that determines the timing of expansion and contraction (Fig. 2c and Fig. 2d).

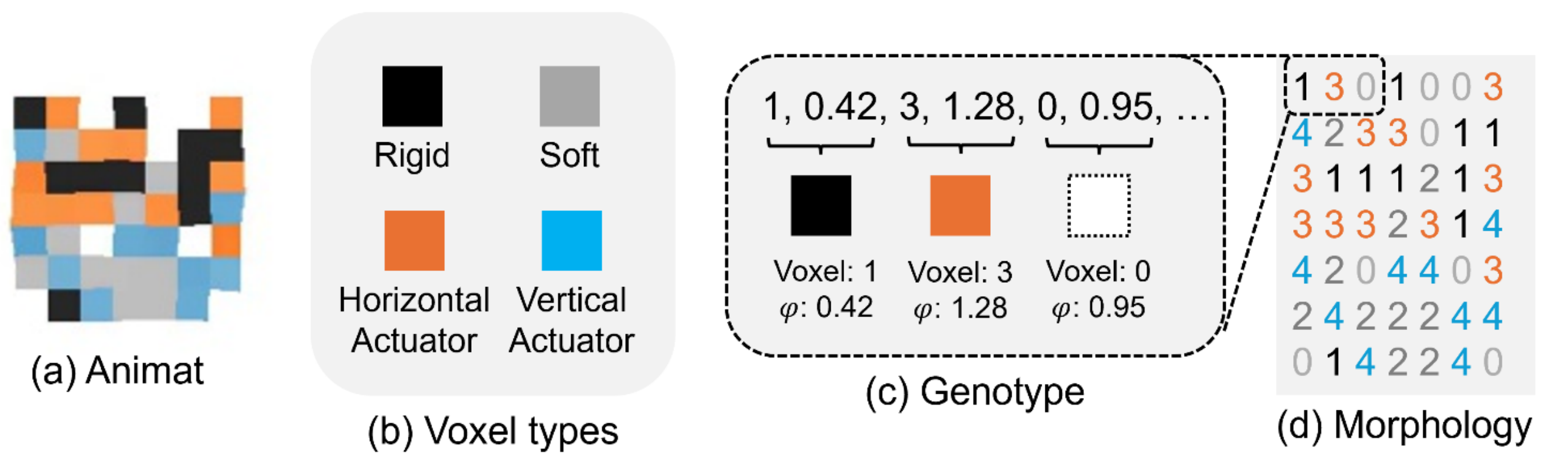

**Fig. 2** Structure and representation of animats. (a) Example of an animat. (b) Types of voxels composing an animat. (c) Genotype of an animat. (d) Morphology of an animat corresponding to the genotype

### 2.3 Selection with VLM-based subjective evaluation

As shown in Fig. 1b, four snapshot images of an animat at predetermined time points are arranged horizontally in chronological order from left to right (Fig. 1c). To eliminate the influence of voxel-specific colors on the evaluation, the body of each animat is rendered uniformly in red. Two such sequences, corresponding to two individuals to be compared, are then stacked vertically to form a single image (hereafter referred to as a sequence image). This sequence image is provided to a VLM together with a prompt (see Appendix), and the model is asked to determine whether the individual in the top or bottom row exhibits motion that better satisfies the evaluation criterion. At the same time, for the purpose of post-hoc analysis, the model is instructed to provide the evaluation criterion underlying its judgment (Fig. 1d).

The prompt is designed to be executable even on small-scale language models that can run in local computing environments, referred to as small language models (SLMs), while still enabling consistent evaluations. Specifically, the prompt first instructs the model to assume the role of an impartial motion analysis specialist ("You are an impartial motion analysis specialist who evaluates motion patterns based solely on their intrinsic qualities"). Next, a description of the input image is provided ("In this image, there are two frame sequences (top and bottom rows), each containing 4 chronological frames (time steps: t, t+1, t+2, t+3) of tracked red objects from left to right"). The model is asked to describe the shape of the animat in each frame in temporal order ("Describe the shape in each frame (t, t+1, t+2, t+3)"), followed by an explanation of how the shape transforms between frames ("Describe how the shape transforms between frames"). Subsequently, the model is prompted to articulate the definition of the subjective evaluation term within the given context (e.g., "Define what '*adorably*' means in the context of motion/transformation"). After completing these steps, the model performs the subjective evaluation based on the specified term. Finally, it is instructed to answer with a single word representing the evaluation criterion and to indicate its selection by responding with either *top* or *bottom*.

The rationale for eliciting the subjective evaluation of motion in a step-by-step manner is to reduce output bias that may arise from presenting explicit choices in the prompt. By encouraging the model to verbalize its reasoning process, the selection can be made more reliably. In addition, to prevent the VLM's decisions from depending on the presentation order of the choices and to exclude ambiguous responses, the same evaluation is performed again with the vertical positions of the two individuals in the input image swapped. The selection result is accepted only when the model makes the opposite positional choice after swapping (i.e., when it consistently selects the same individual).

### 2.4 Evolution

As described in Section 2.1, the animats evolve through a genetic algorithm. A population of $N$ individuals evolves iteratively until the number of generations reaches $G$. Individuals in the initial generation are generated at random. In each generation, two individuals are randomly sampled from the population and evaluated via the pairwise comparison method described in Section 2.3, with the winner designated as a parent individual. This procedure is repeated until the number of parent individuals equals the population size. Crossover is applied with probability $p_{cross}$. When crossover occurs, one-point crossover is performed between the two parent individuals, with the crossover position chosen at random. Mutation is applied independently to each locus with probability $p_{mut}$.

When a mutation occurs, the value at the corresponding locus is randomly replaced with an alternative admissible value. Due to the constraints imposed by the Evolution Gym framework, every voxel composing an individual must share at least one edge with another voxel. If crossover or mutation produces an individual that violates this constraint, a repair procedure is applied: a locus is randomly selected and its value is changed to another random value, and this process is repeated until the constraint is satisfied.

## 3 EXPERIMENTS

### 3.1 Experimental settings

The experiments were conducted with a population size of $N = 30$, a total of $G = 50$ generations, a crossover probability of $p_{cross} = 0.05$, and a mutation probability of $p_{mut} = 0.01$. The maximum height and width of an animat were set to $W = 7$, and each evaluation was performed over $t = 600$ time steps. As the VLM, we used a 4-bit quantized version of gemma-3-12b-it[1][14], with the temperature parameter set to 0. All inference was performed on an NVIDIA GeForce RTX 4080 SUPER.

For subjective evaluation in the evolutionary process, we conducted experiments under five different conditions using evaluation terms with distinct semantic orientations: *adorably*, *weirdly*, *solemnly*, *dynamically*, and *motionlessly*. The terms *adorably*, *weirdly*, and *solemnly* were selected as highly abstract expressions that allow for diverse interpretations, whereas *dynamically* and *motionlessly* were chosen as expressions that more directly evoke specific modes of motion.

### 3.2 Validity of VLM-based subjective selection

It is not self-evident to what extent selection by a VLM, based on subjective linguistic expressions that are the focus of this study, induces a directional bias or selective pressure in evolution. To address this issue, we first quantified intra-population diversity and compared the results with those obtained under random selection. Specifically, for each generation, we computed the Hamming distance between the genotype of each individual and those of all other individuals in the same generation. We then averaged these distances for each individual and further averaged the resulting values across the population to obtain the mean Hamming distance within the generation.

Figure 3 shows the temporal evolution of the mean Hamming distance for each subjective evaluation term. The dotted lines represent individual trials, and the solid black line indicates the average over five trials. Focusing on the five-trial averages, when evolution was driven by random selection, the mean Hamming distance gradually decreased at an approximately constant rate, reaching 33.9 in the final generation. This indicates that population convergence occurs at a steady pace. In contrast, under all other conditions, the mean Hamming distance decreased sharply from the early generations compared to the random selection case and fell below 20 in the final generation for all conditions. These results indicate that the subjective evaluations provided by the VLM led to stronger selection of individuals possessing certain characteristics.

Because the selection method in the proposed model does not compute an absolute fitness value but instead determines relative winners through pairwise comparisons, it is not possible to quantitatively track adaptive progress across generations in a conventional manner. Therefore, to evaluate whether the animats evolved toward morphologies and locomotion more consistent with the subjective evaluation terms, we conducted post-hoc pairwise comparisons between individuals from the initial and final generations. In each comparison, individuals from the initial and final generations were randomly paired such that each individual from both generations was evaluated an equal number of times. Pairwise comparisons were then performed for each pair in the same manner as during the evolutionary process.

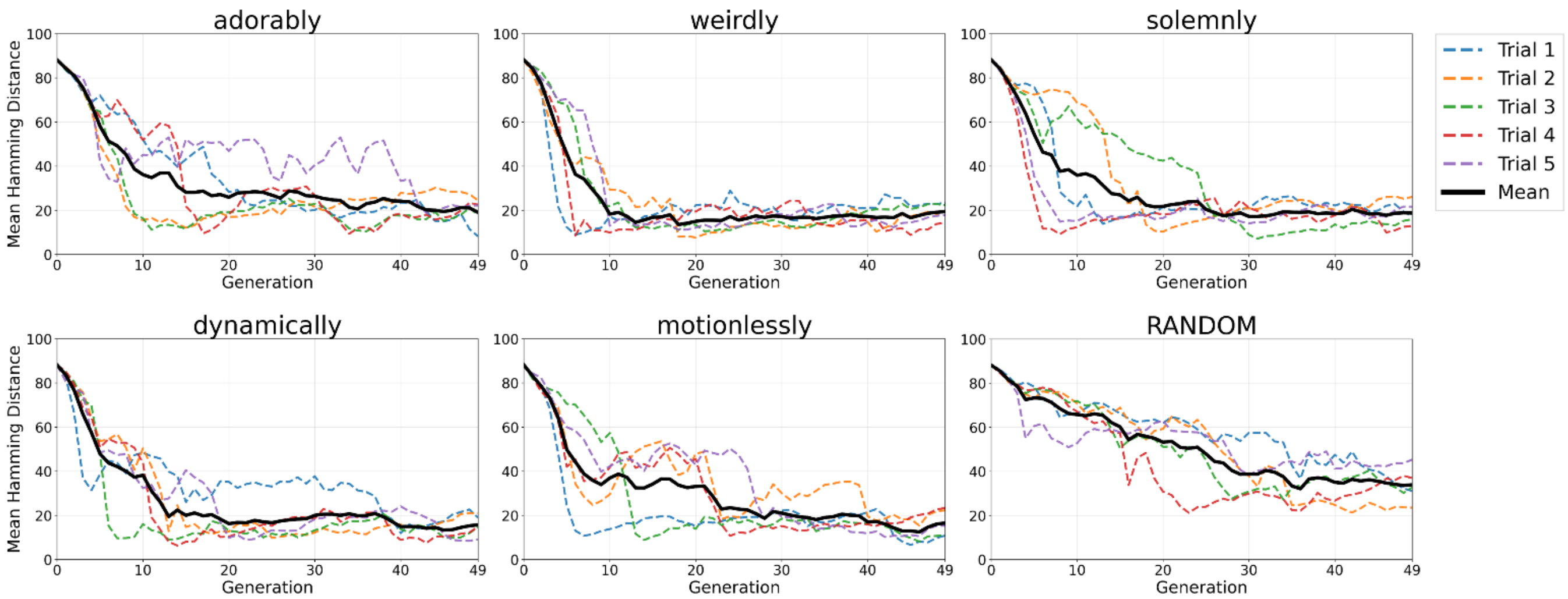

**Fig. 3** Temporal changes in the mean Hamming distance between individual genotypes for each subjective evaluation term. Dotted lines indicate values for individual trials, and the solid black line represents the average over five trials

[1] https://huggingface.co/lmstudio-community/gemma-3-12b-it-GGUF

Figure 4 shows, for each subjective evaluation term, the average number of times individuals from the initial generation and the final generation were selected, as well as the number of ties, along with 95% confidence intervals over five trials. Here, Draw indicates that the model's responses were inconsistent when the vertical positions of the two individuals in the input image were swapped. For all subjective evaluation terms, individuals from the final generation were selected more frequently than those from the initial generation. This suggests that, because of evolution, the animats adapted toward morphologies and motions that better aligned with the given subjective evaluation terms under each condition.

However, the magnitude of this effect varied depending on the evaluation term. Specifically, for *adorably* and *solemnly*, the number of ties was close to the number of selections favoring final-generation individuals, and the difference between the initial and final generations was relatively small. This suggests that, for these evaluation terms, the VLM's judgments were less decisive for some pairs, and that evolutionary differences may not have manifested as clearly distinguishable visual features. In contrast, for *weirdly*, *dynamically*, and *motionlessly*, the number of ties was relatively small, and the number of selections favoring final-generation individuals increased substantially. These results suggest that, for these evaluation terms, evolution led to morphologies and locomotion that were more readily distinguishable by the VLM in visual terms. In other words, these evaluation terms are likely more strongly correlated with visually salient morphological or motion features.

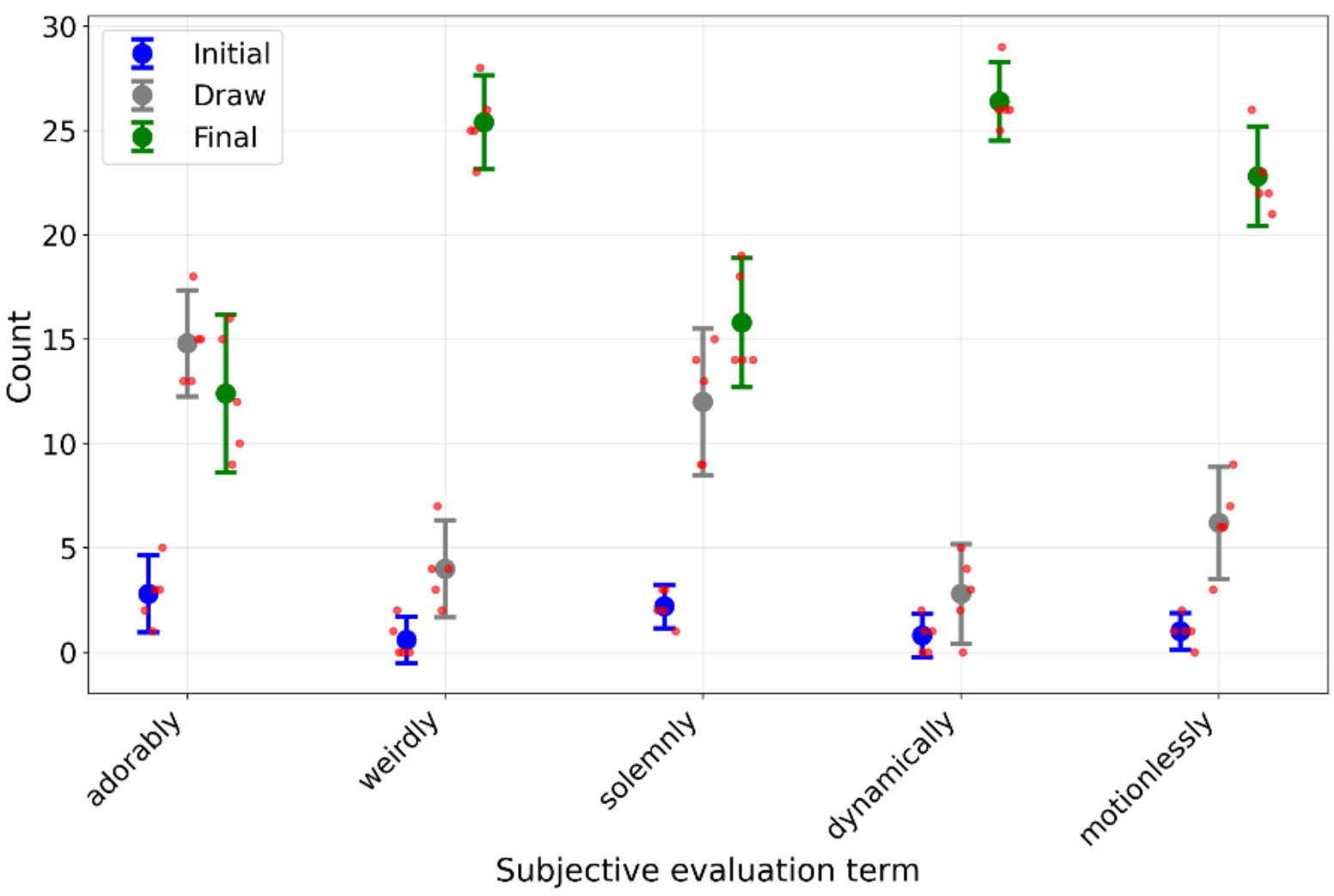

**Fig. 4** Mean number of selections in pairwise comparisons between the initial and final generation populations, averaged over five trials. Error bars indicate 95% confidence intervals, and red points represent the values from individual trials

### 3.3 Evolved animats

Figure 5 presents sequence images of, for each trial, the individual in the final generation whose genotype exhibited the minimum average Hamming distance to the genotypes of all other individuals (i.e., the individual whose genotype was closest to the population center in the final-generation population of that trial). Figure 6, in turn, shows a two-dimensional projection of sequence images from all individuals across all conditions and trials. Specifically, each sequence image was encoded as a high-dimensional feature vector using CLIP [15], and these vectors were projected into a two-dimensional space using UMAP [16] (n_neighbors = 50, min_dist = 0.5). Only individuals from the final generation are plotted in the figure.

To further analyze the morphological characteristics of animats evolved under subjective evaluation terms as fitness criteria, we evaluated how each subjective term influenced the number of voxels composing the animats. For all individuals in the final generation of each trial, the number of voxels was counted by voxel type. Subsequently, the average number of voxels per individual was computed within each trial. Figure 7 shows the mean voxel counts for each subjective evaluation term, where points represent the averages over five trials and error bars indicate 95% confidence intervals.

Under the *adorably* condition, two distinct types of individuals were observed: those with leg-like structures exhibiting walking-like motions (Trials 1, 2, and 5 of Fig. 5), and those with more blob-like morphologies that deformed in a bouncing manner (Trials 3 and 4 of Fig. 5). The former can be interpreted as expressing adorableness through small-animal-like locomotion, whereas the latter correspond to adorableness derived from a soft and flexible texture. These results suggest that different notions of *adorableness* emerged across trials. In the UMAP space, clusters formed in different regions depending on the trial (central, lower, and left regions in Fig. 6), indicating that evolution under the *adorably* condition proceeded in different directions across trials.

Under the *weirdly* condition, slit-like structures and motions resembling opening, closing, and expansion of these slits were observed (Fig. 5). Irregular deformations resembling the body splitting apart can be interpreted as producing an impression of *weirdness*. Although the *dynamically* condition was selected as a term that directly evokes a mode of motion, large body-expanding movements were also commonly observed, similar to the *weirdly* condition. This suggests that the VLM interpreted *dynamic* primarily in terms of large-amplitude motion or rapid deformation. In the UMAP space, relatively isolated clusters were formed for each trial under both the *weirdly* and *dynamically* conditions. This indicates that, although these conditions shared common motion patterns (e.g., opening and closing of slit-like structures), the directions and specific realizations of motion differed depending on morphology. Focusing on voxel composition, both the *weirdly* and *dynamically* conditions exhibited the highest average number of vertical actuator voxels (Fig. 7), suggesting that large body-expanding motions may be associated with an increase in vertical actuators.

Under the *solemnly* condition, individuals with near-square outlines and internal hollow structures were observed. Stable shapes and gradual deformations can be interpreted as corresponding to an impression of *solemnity*. Although the *motionlessly* condition was chosen as a term that directly evokes a mode of motion, many individuals also exhibited near-square morphologies, converging toward phenotypes similar to those under the *solemnly* condition. In the UMAP space, there were trials in which individuals from both conditions converged into the same region (e.g., Trial 2 in the upper region of Fig. 6), suggesting that the VLM evaluated *motionlessness* and *solemnity* using similar criteria. Regarding voxel composition, the *solemnly* condition exhibited the highest average number of soft voxels, whereas the *motionlessly* condition exhibited the highest average number of rigid voxels (Fig. 7). Overall, both conditions tended to have a larger number of static voxels, suggesting that gradual or minimal deformation may correspond to an increased proportion of static voxels.

Taken together, these results confirm that, through VLM-based selection, morphologies and locomotion reflecting the characteristics of each subjective evaluation term can evolve. However, even among abstract evaluation terms, differences were observed: some terms, such as *adorably*, led to divergence into different phenotypes across trials, whereas others, such as *weirdly* and *solemnly*, exhibited relatively consistent evolutionary directions. Furthermore, the terms *dynamically* and *motionlessly*, which were selected as expressions that directly evoke modes of motion, converged toward phenotypes similar to those evolved under *weirdly* and *solemnly*, respectively. These findings suggest that the VLM does not apply evaluation terms in a purely literal

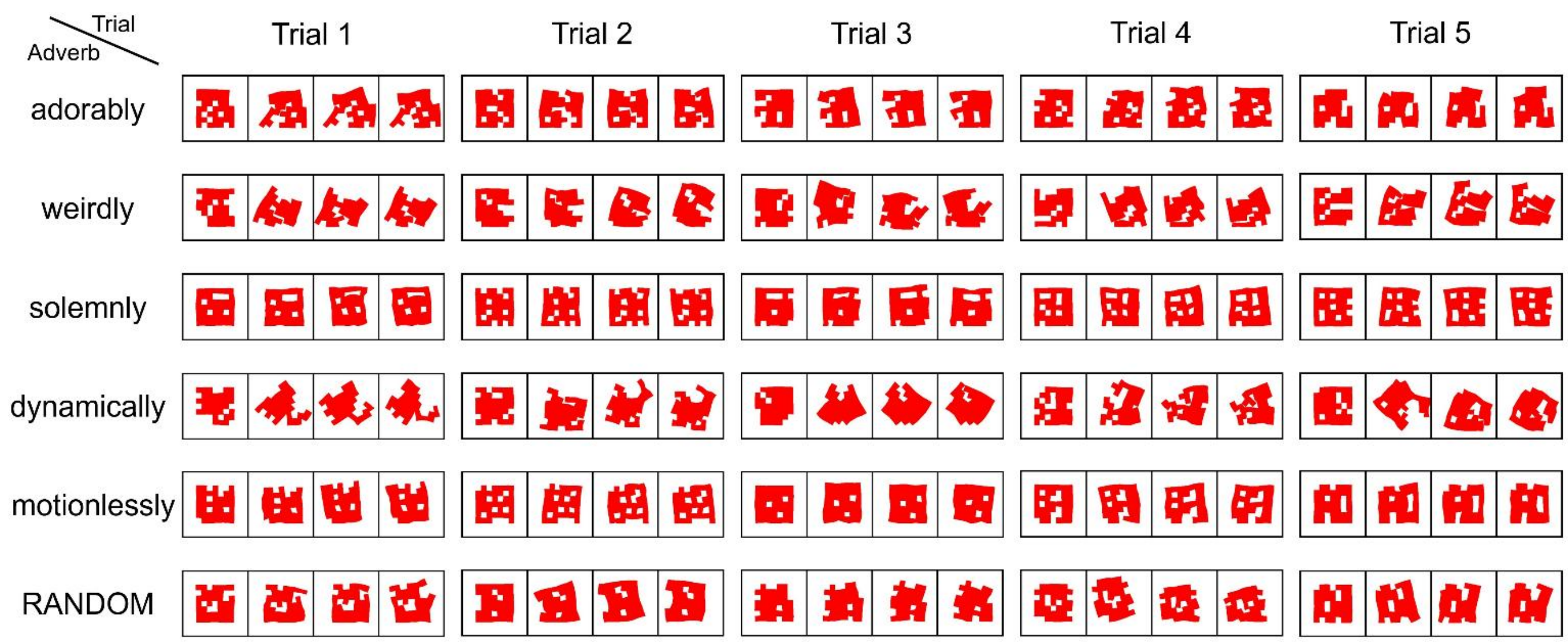

**Fig. 5** Sequence images of representative final-generation individuals for each trial and subjective evaluation term. For each condition and trial, the individual shown is the one in the final generation with the minimum average Ham ming distance between its genotype and those of all other individuals

manner, but instead decomposes them into internal evaluation criteria, and that similarities among these criteria determine similarities in evolutionary outcomes.

The interpretations of morphology and motion presented in this section should be understood as one possible perspective on the relationship between subjective evaluation terms and visual features, rather than as definitive mappings. The same phenotype may be associated with different subjective meanings depending on the observer or analytical viewpoint. Therefore, the qualitative descriptions provided here are intended to support the interpretation of the relationship between the evolved phenotypes and the VLM-derived evaluation criteria, but they do not exclude alternative interpretations.

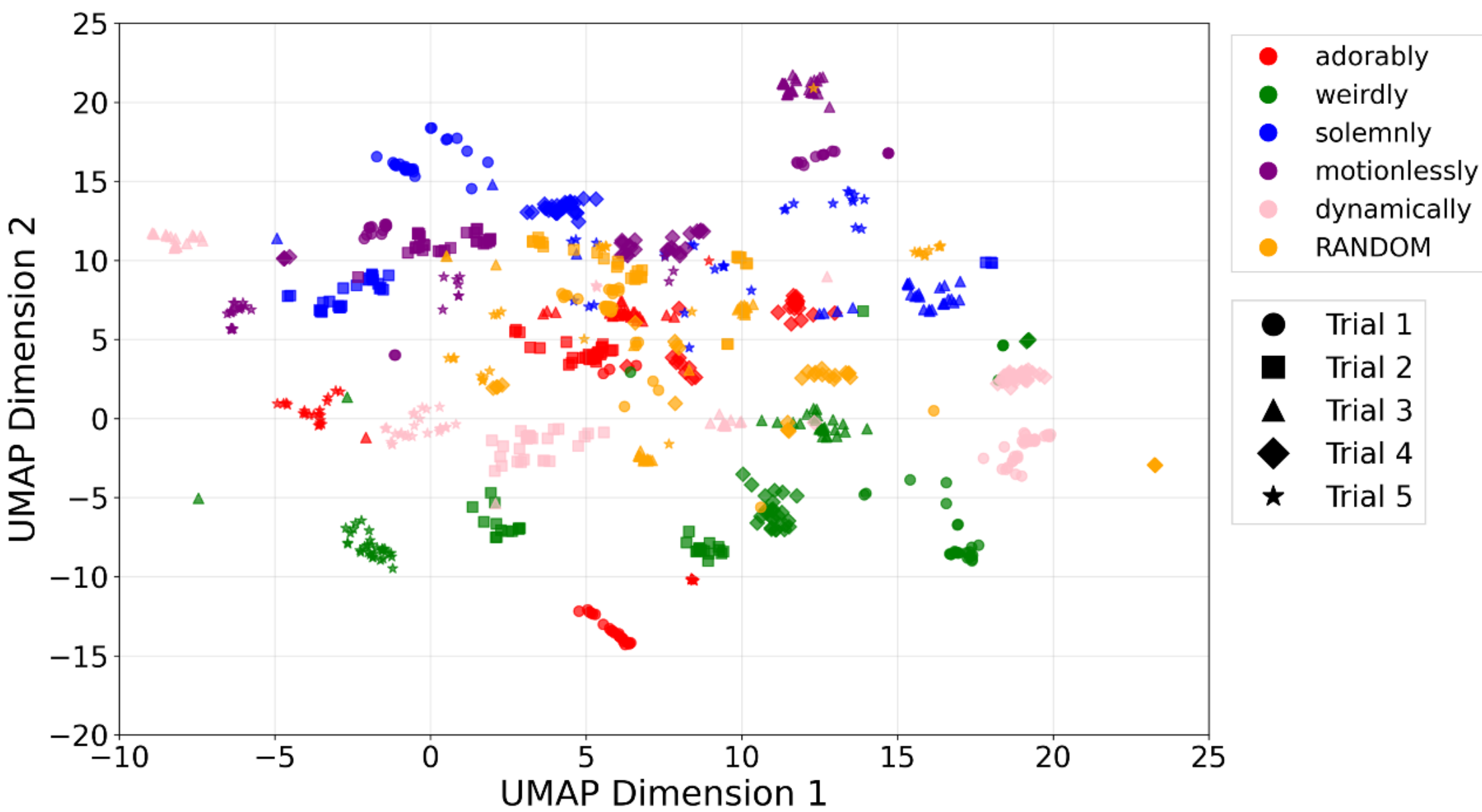

**Fig. 6** Two-dimensional projection of final generation animats in phenotype space using UMAP

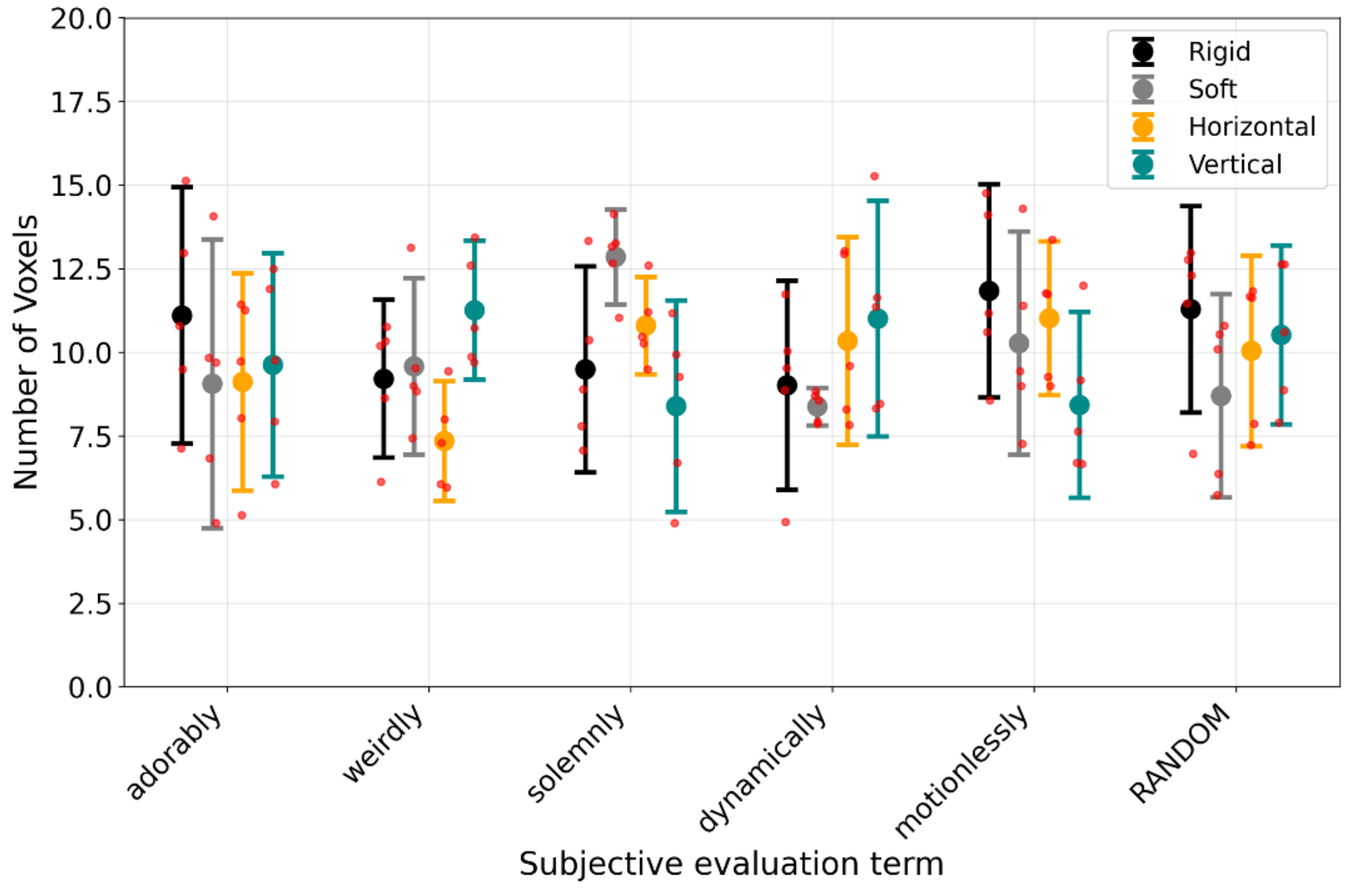

**Fig. 7** Mean number of voxels composing final generation animats, averaged over five trials. Error bars indicate 95% confidence intervals, and red points represent values from individual trials

### 3.4 Evaluation criteria

To investigate the decision tendencies of the VLM when performing pairwise comparisons based on subjective evaluations, we aggregated the evaluation criteria terms output during comparisons across all generations and all individuals in all five trials. For each subjective evaluation term, we then computed the relative frequency of the evaluation criteria terms. Table 1 lists, for each subjective evaluation term, the top five most frequently occurring evaluation criteria along with their occurrence ratios.

Furthermore, to quantitatively assess the similarity of evaluation criteria across different subjective evaluation terms, we computed the Jaccard indices between sets of evaluation criteria. Specifically, for each subjective evaluation term, we constructed a set of evaluation criteria that included all criteria appearing at least once across the five trials. Then, for every pair of subjective evaluation terms, the Jaccard indices were calculated based on the intersection and union of the corresponding evaluation-criteria sets. Figure 8 presents a heatmap of the Jaccard indices for all pairs of subjective evaluation terms. Larger values indicate that the VLM relied on more similar evaluation criteria when making judgments for the two subjective terms.

For *adorably*, the evaluation criteria exhibited a strong bias toward the term *fluidity*. Moreover, the top three evaluation criteria accounted for more than 80% of all occurrences (Table 1), indicating that the criteria used for *adorably* were more consistent than those for other subjective evaluation terms. Examining the Jaccard indices with other terms, the indices with *weirdly* was 0.104, while those with the remaining terms were approximately 0.2 (Fig. 8).

For *weirdly*, the most frequently occurring evaluation criterion was *disjunction*. Among the Jaccard indices with other subjective evaluation terms, the indices with *dynamically* was the highest at 0.192. Consistent with this result, the top five most frequent criteria for both *weirdly* and *dynamically* included *abruptness* and *unpredictability*. This suggests that large shape changes and irregular deformations were associated with unpredictability, potentially leading to the evolution of morphologies featuring slit-like structures. This result is consistent with the similarities observed in the morphologies and locomotion of final-generation individuals discussed in Section 3.2.

For *solemnly*, the most frequent evaluation criterion was *deliberation*. Among the Jaccard indices with other subjective evaluation terms, the indices with *motionlessly* were the highest at 0.312. In addition, the top five most frequent criteria for both *solemnly* and *motionlessly* included *regularity* and *continuity*. This suggests that stable, near-square morphologies and regular motion patterns were associated with these criteria and reflected in the selection pressure. This result also aligns with the similarity in morphologies and motions of final-generation individuals reported in Section 3.2.

Overall, a clear correspondence was observed between the similarity of evolved morphologies and the frequency patterns of the evaluation criteria. These findings support the interpretation suggested in the previous section: the internal criteria employed by the VLM during judgment are externalized as evaluation criteria, and the similarity of these internal criteria is reflected in the similarity of evolutionary outcomes.

**Table 1** Top five most frequent evaluation criteria terms used for each subjective evaluation, along with their occurrence ratios (rounded to three decimal places)

| rank | adorably | weirdly | solemnly | dynamically | motionlessly |
|---|---|---|---|---|---|
| 1 | Fluidity | Disjunction | Deliberation | Unpredictability | Regularity |
| | 0.461 | 0.175 | 0.241 | 0.200 | 0.251 |
| 2 | Whimsy | Abruptness | Gradualism | Irregularity | Continuity |
| | 0.194 | 0.172 | 0.166 | 0.147 | 0.129 |
| 3 | Gentleness | Discontinuity | Regularity | Variance | Uniformity |
| | 0.152 | 0.157 | 0.076 | 0.139 | 0.077 |
| 4 | Playfulness | Unpredictability | Continuity | Fluency | Stability |
| | 0.030 | 0.155 | 0.074 | 0.117 | 0.066 |
| 5 | Rhythm | Fragmentation | Predictability | Abruptness | Subtlety |
| | 0.026 | 0.054 | 0.058 | 0.117 | 0.047 |

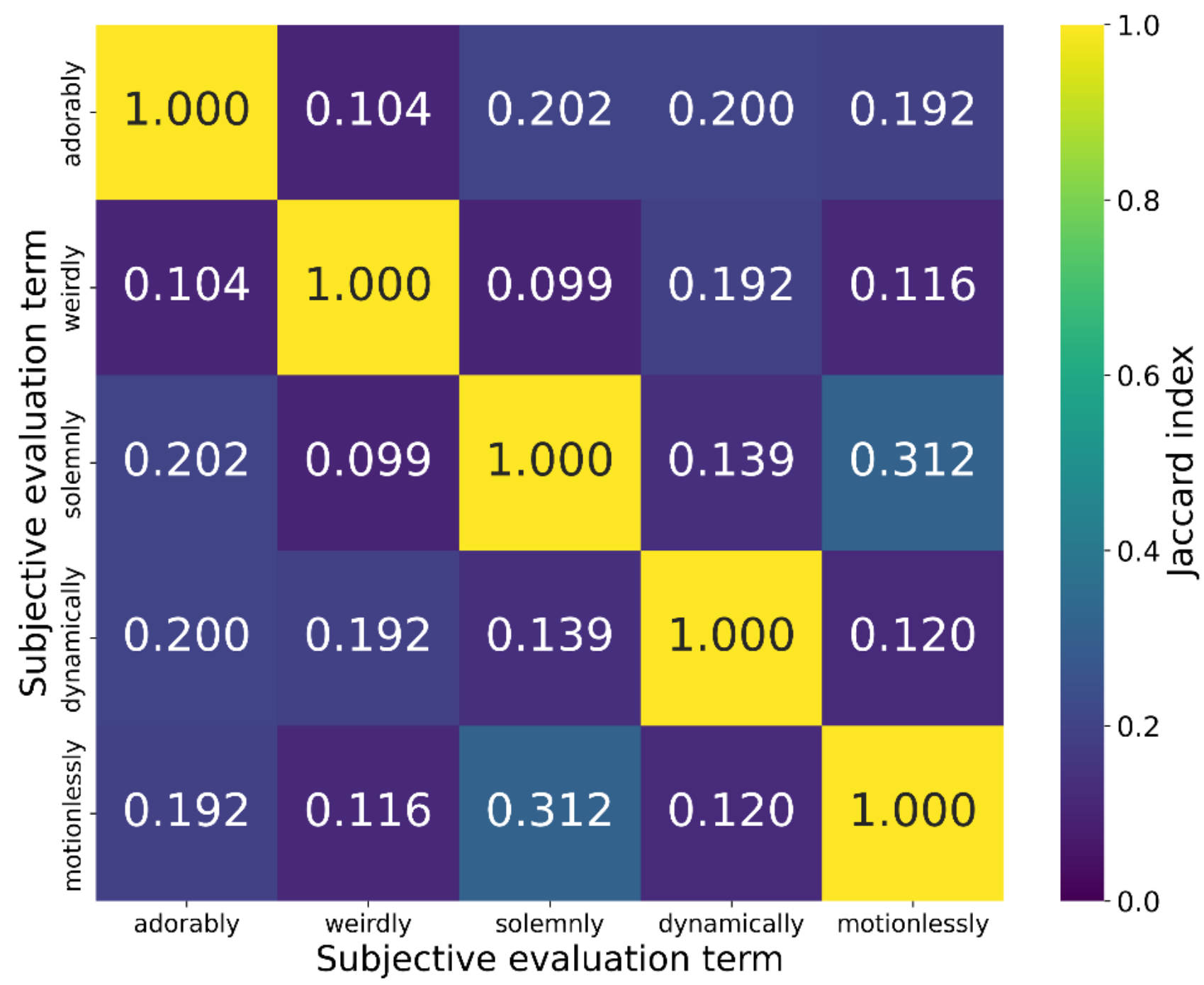

**Fig. 8** Jaccard indices of evaluation criteria for each pair of subjective evaluation terms

### 3.5 Comparison with human subjective evaluation

To examine the extent to which the VLM-based subjective evaluation proposed in this study corresponds to human subjective perception, we additionally conducted an interactive evolutionary computation experiment with human participants. However, the primary objective of this study is not to achieve complete agreement with human evaluations, but rather to investigate how evolution driven by VLM-based subjective selection affects selection pressure and phenotypes. Therefore, in this section, we do not regard human evaluations as an absolute ground truth. Instead, we position this experiment as an auxiliary comparison for examining the extent to which the evolutionary outcomes induced by the VLM overlap with, and differ from, human subjective evaluations. We also administered a questionnaire to assess the practical burden of human-subjective IEC, including fatigue, perceived difficulty, and impressions of the evolutionary process.

#### 3.5.1 Human-subjective IEC setup

The experiment with human participants was conducted with nine undergraduate and graduate students who were native speakers of Japanese. The experiment was conducted in Japanese. The participants consisted of five males and four females, and their mean age was 21.67 years. All experimental procedures were approved by the research ethics review of the Graduate School of Informatics, Nagoya University (approval number: I26-04). All participants provided written informed consent prior to participation.

Among the evaluation terms used in the VLM experiments, we focused on three conditions, namely *adorably*, *weirdly*, and *solemnly*, which are highly abstract expressions that allow for diverse interpretations. For each of the three conditions, three trials with different initial populations were conducted, resulting in nine trials in total. Each trial was evaluated by a different participant, so that each participant evaluated one trial only. This setting was adopted to secure the minimum set of conditions necessary for comparing VLM-based evaluation with human evaluation, while considering the time cost and evaluator fatigue associated with human evaluation.

The number of selection generations was set to 20 because, in the VLM experiments, the mean Hamming distance among genotypes within the population tended to converge around generation 20 (Fig. 3). This setting also accounts for evaluator fatigue, which cannot be ignored in IEC because participants must repeatedly perform pairwise comparisons among many individuals.

In each trial, we used the same combinations of initial populations and tournament selections as in the VLM experiments. The participants were informed that the top and bottom rows of each input image were frame-by-frame images showing the motion of virtual organisms arranged chronologically from left to right. They were then asked to select the individual that they felt better matched the specified evaluation term, considering not only its morphology but also the motion inferred from the image sequence. After each selection, the participants were asked to freely describe the features they considered important in making their judgment. Figure 9 shows the experimental interface used in the human-subjective IEC experiment. In each selection step, the interface presented the evaluation term, the two candidate sequence images arranged in the top and bottom rows, response options for selecting one of the candidates, and a text box for freely describing the features emphasized in the judgment.

**Fig. 9** Example of the experimental interface presented to the participants. The instructions are shown at the top, and the images used for selection are displayed on the left. Participants selected an individual using the buttons next to the images. They also entered free-text descriptions of the features emphasized during evaluation in the text box at the lower right. History buttons were provided above the text box to allow previously entered descriptions to be reused easily

### 3.5.2 Comparison between VLM-based and human-based evolution

First, we compared the VLM-based and human-based evaluations at the level of individual pairwise selections in the initial generation. To make this comparison direct, we used the same initial populations and the same tournament-selection pairs in the human-subjective IEC experiment as in the VLM-based experiment. For each tournament pair, we recorded whether the VLM and the human participant selected the same individual and calculated the agreement rate over the 30 tournament selections under each condition. This comparison provides an estimate of how closely the VLM, and human participants agreed at the outset, before evolutionary dynamics affected the effects of their selection tendencies.

Table 2 shows the agreement rates for the initial-generation selections. When 50% is taken as the chance level, the agreement rate was below chance for *adorably* (43.3%), suggesting that the VLM and human participants did not necessarily rely on the same criteria for judging adorableness. By contrast, the agreement rates were modestly above chance for *weirdly* (62.2%) and *solemnly* (55.6%). These results suggest that agreement between the VLM and human participants was limited at the level of individual pairwise choices but varied depending on the evaluation term: weak directional agreement was observed for *weirdly* and *solemnly*, whereas *adorably* showed a larger gap. This limited agreement should be interpreted with caution, however, especially because the task was highly subjective. For such subjective tasks, the low agreement between the choices made by the VLM and those made by human participants may arise not only from inter-participant variability but also from within-participant inconsistency. The additional experiment was too limited to examine this possibility.

**Table 2** Three-trial average agreement rates between VLM and human choices in tournament selection in the initial generation under each condition

| | Trial 1 | Trial 2 | Trial 3 | Mean |
|---|---|---|---|---|
| adorably | 0.433 | 0.533 | 0.333 | 0.433 |
| weirdly | 0.767 | 0.433 | 0.667 | 0.622 |
| solemnly | 0.500 | 0.567 | 0.600 | 0.556 |

Next, we compared the population-level evolutionary dynamics induced by the two evaluation schemes by analyzing temporal changes in the mean Hamming distance within the population. For each generation, we computed the Hamming distance between the genotype of each individual and the genotypes of all other individuals in the same population. These distances were first averaged for each individual and then averaged across all individuals to obtain the mean Hamming distance for that generation. A lower mean Hamming distance indicates that the genotypes in the population have become more similar, suggesting stronger convergence.

Figure 10 shows the mean Hamming distance in each generation under human-based and VLM-based evaluation. Under both schemes, the mean Hamming distance decreased over generations while showing fluctuations, and by the 20th generation it reached a comparable level, with no substantial difference between the two schemes. These results indicate that human evaluation, like VLM evaluation, imposed a certain degree of selection pressure based on abstract subjective evaluation terms and evolved the population in specific directions. For *adorably* and *weirdly*, human evaluation appeared to converge somewhat more slowly and with larger fluctuations than VLM evaluation. This might reflect a lower consistency in human judgments, although the small number of trials precludes any firm conclusion. This tendency was not observed under *solemnly*.

Next, we compared the individuals and their motions in the final generation. Figure 11 shows the representative individual (defined as in Fig. 5) in the 20th generation under each condition.

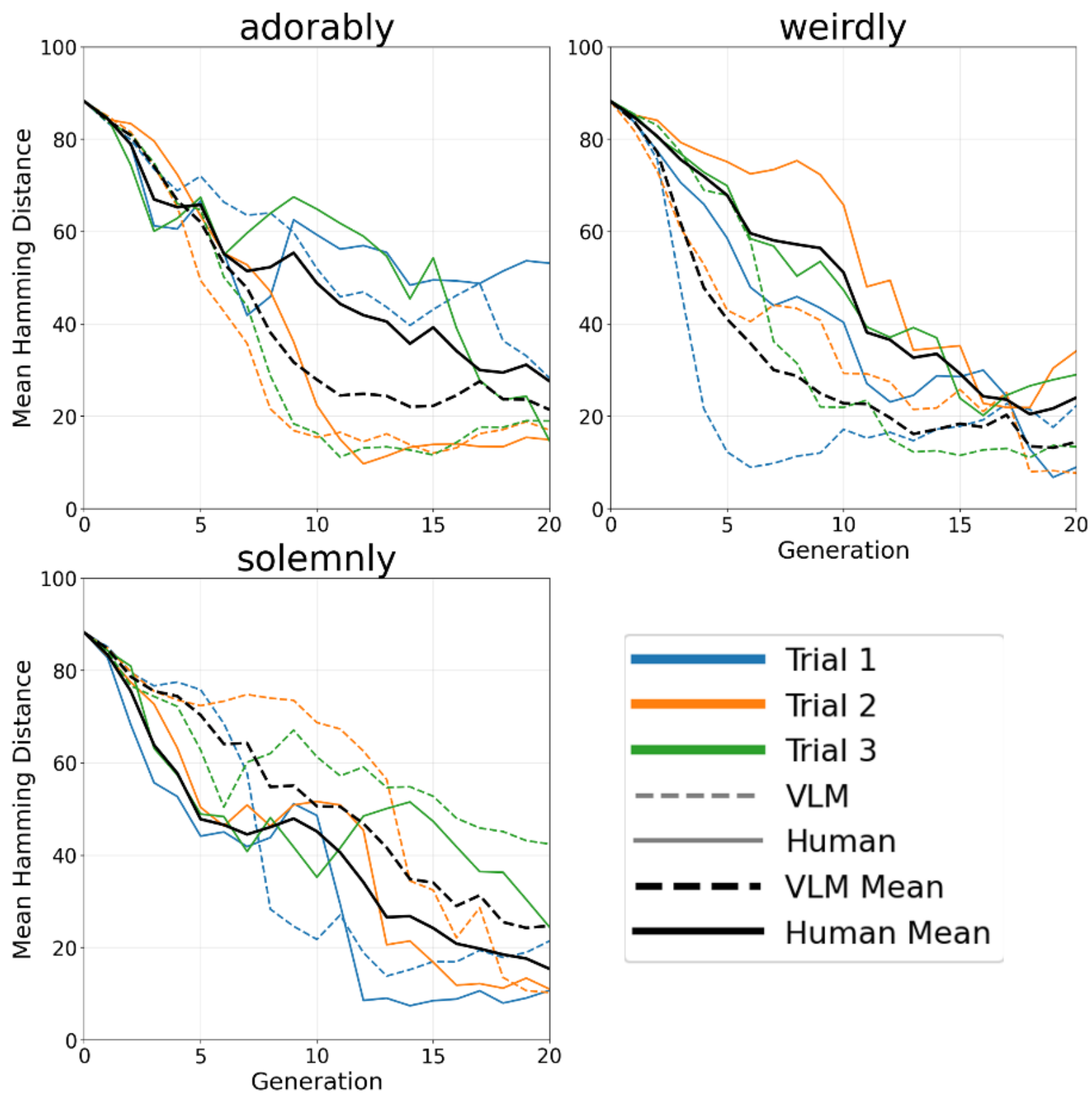

**Fig. 10** Comparison of changes in genotypes across generations. Solid lines indicate the results of the experiments based on human evaluation, whereas dotted lines indicate the results of the experiments based on VLM evaluation. Each color corresponds to a trial, and black indicates the average over the three trials

Under the *adorably* condition, in the experiments based on human evaluation, the animats evolved into morphologies and motions resembling animal eyes or bodies (Fig. 11, Trials 2 and 3). In the free-description responses regarding the features emphasized during judgment, human participants mentioned an appearance of joy (Trial 1), shape transformations resembling an animal face (Trial 2), and small-animal-like movements (Trial 3). These responses suggest that human participants tended to associate adorableness with animal-like morphologies and structures suggestive of eyes or bodies. This tendency can be interpreted as partially similar to the leg-like structures observed in the VLM-based experiments. However, because human participants associate adorableness more explicitly with animal-like features, this tendency appeared more clearly in human-based experiments.

Under the *weirdly* condition, in the experiments based on human evaluation, the animats evolved motions in which the body spread widely (Fig. 11, Trials 1, 2, and 3). In the free-description responses, participants tended to focus on geometric changes in shape (Trial 1), changes in motion (Trial 2), and large changes in angle (Trial 3). These tendencies were qualitatively similar to the slit-like structures and opening-and-closing motions observed in the VLM-based experiments. This suggests that, among the three conditions examined in this experiment, *weirdly* was the evaluation term for which the criteria of the VLM and human participants overlapped relatively well. In Trial 3 of the human-based experiment, the animat appears smaller from the third frame onward. This occurred because, in the Evolution Gym setting, when the animat moved outside the display area, the view was scaled down so that the entire body remained visible.

Under the *solemnly* condition, in the experiments based on human evaluation, the animats evolved near-square morphologies with few empty spaces and high density, accompanied by only slight deformation. In the free-description responses, participants tended to focus on the small amount of motion (Trial 1) and the lack of empty space (Trials 2 and 3). In the VLM-based experiments, near-square outlines were also observed under the *solemnly* condition, indicating similarity with the human-based experiments in this respect. On the other hand, whereas the VLM-based experiments often produced morphologies with internal hollow structures, human participants tended to emphasize the lack of empty space. This difference suggests that the VLM may have placed greater emphasis on stable contours and regular deformation, whereas human participants may have placed greater emphasis on density and a sense of heaviness.

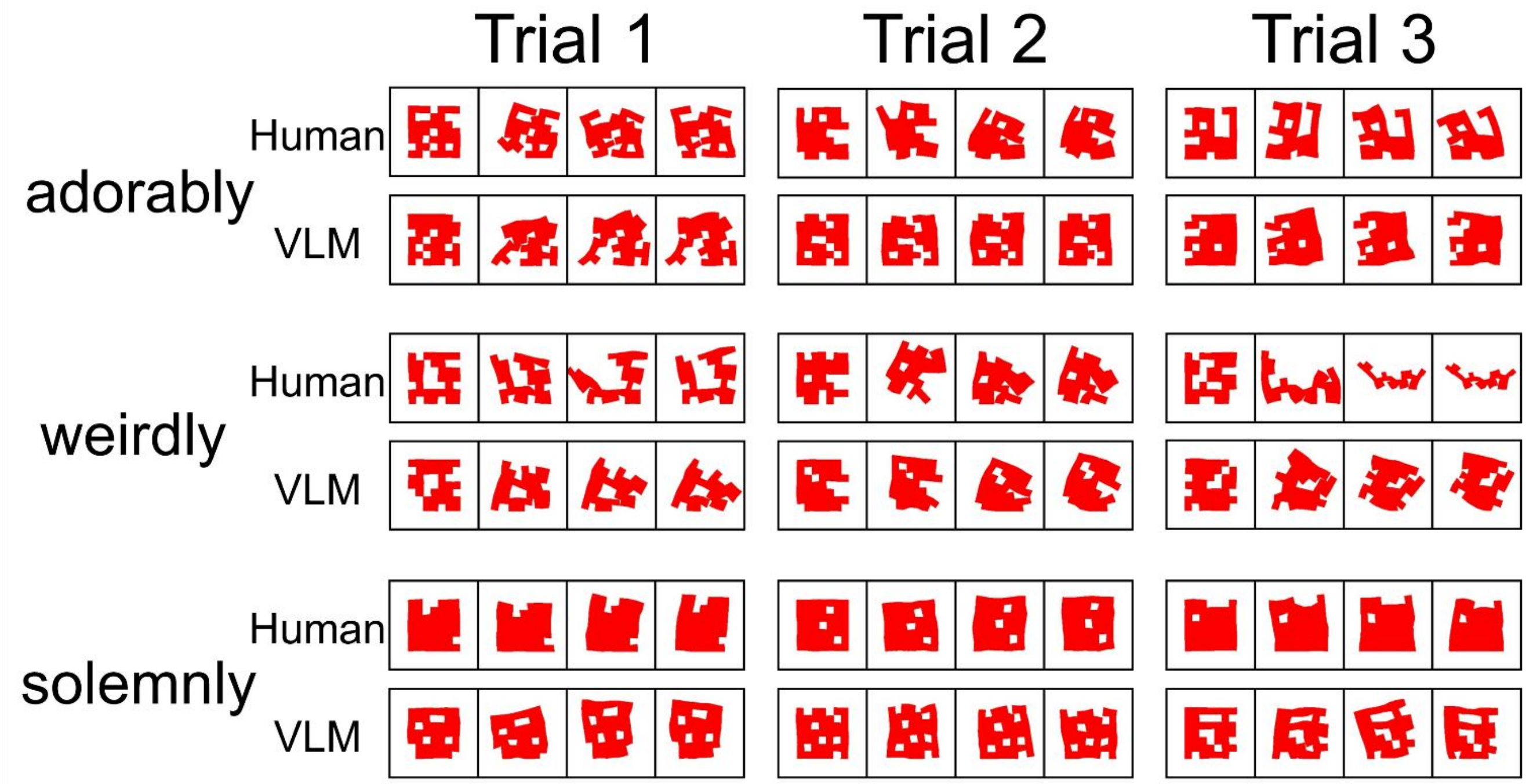

**Fig. 11** Sequence images of representative individuals in the final generation (20th generation) for each trial and each subjective evaluation term. For each condition and each trial, the individual shown is the one whose genotype had the minimum mean Hamming distance to the genotypes of all other individuals in the final generation

Furthermore, to visualize the relationships among the characteristics of the evolved populations, the sequence images of the final-generation individuals in each condition and trial were input to gemma-3-12b-it. The model was prompted to describe each individual based on the prompt shown in the Appendix. The resulting descriptions were then vectorized using Qwen3-Embedding-0.6B[2], and the obtained vectors were projected into a two-dimensional space using UMAP [16] (n_neighbors = 50, min_dist = 0.5).

Figure 12 shows the results. Under the *adorably* condition, individuals with leg-like structures were observed in both human-based and VLM-based experiments. However, the specific morphologies were diverse, and the points were distributed over a wide area. This indicates that animal-like adorableness is not based on a single visual feature but can be realized through multiple different features. Under the *weirdly* condition, although the morphologies themselves varied across trials, they shared the common feature of large deformation associated with motion. As a result, the human-based and VLM-based results tended to be located in relatively close regions. Under the *solemnly* condition, because the animats generally evolved toward near-square morphologies, the human-based and VLM-based results also tended to be located in relatively close regions.

Finally, the questionnaire responses indicated practical aspects of human-subjective IEC. First, the repeated pairwise selection imposed a clear burden on participants: fatigue tended to increase and concentration to decrease after the experiment, and most participants reported feeling burdened by the repeated evaluations, even though they generally found the task itself interesting. Second, the perceived difficulty of evaluation depended on the term, being relatively low for *adorably* and higher for *solemnly*. At the same time, participants tended to feel that individuals more consistent with the target term emerged in later generations, even though they were not told that the experiment involved evolutionary computation. These results suggest that human-subjective IEC can reflect participants' subjective impressions, while evaluator fatigue and term-dependent difficulty remain important practical limitations.

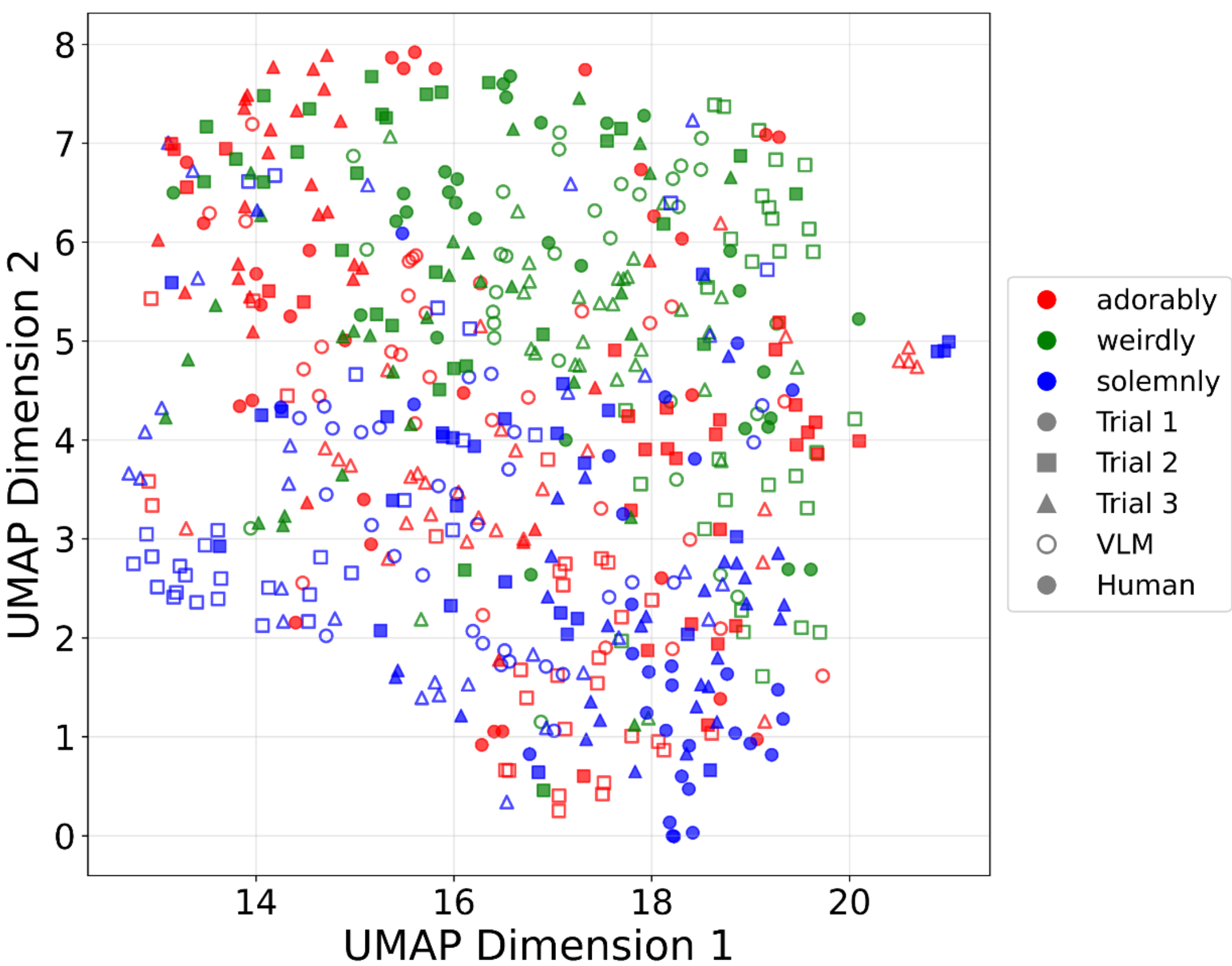

**Fig. 12** Two-dimensional UMAP projection of the features of final-generation individuals from the human evaluation experiment and 20th-generation individuals from the VLM experiment

[2] https://huggingface.co/Qwen/Qwen3-Embedding-0.6B

## 4 CONCLUSIONS

In this study, we proposed an evolutionary computation framework that incorporates subjective evaluations pro-vided by VLM. Results from animat evolution experiments using this framework demonstrated that the animats evolved morphologies and locomotion that reflected the given subjective evaluation terms. The observation that similar morphologies emerged under different subjective evaluation terms suggests that the VLM's subjective judgments are guided by its own internal criteria, and that the evaluation criteria extracted in this study are manifestations of those internal standards.

We also conducted an auxiliary human-subjective IEC experiment to examine the relationship between VLM-based and human-based subjective selection. The results showed that, although humans and VLM did not fully agree at the level of individual selections, the evolutionary outcomes induced by the VLM and those induced by humans exhibited similar tendencies in the morphology and motion of the evolved animats.

In addition, the questionnaire results showed that human-subjective IEC involved fatigue and task burden caused by re-peated pairwise evaluations, despite participants generally perceived the evolutionary process as meaningful. This finding supports the practical motivation for automating subjective evaluation using VLMs.

A limitation of this study is that motion was evaluated using four static snapshots rather than full video input. Although this representation reduces the computational cost of repeated VLM inference, it compresses temporal information and does not explicitly preserve intermediate trajectories between frames. Therefore, rhythm, velocity changes, acceleration changes, and the distinction between abrupt and gradual deformation may not be fully captured. As a future direction, it would be valuable to compare the present sequence-image representation with full-video input or denser temporal sampling, which could reveal how different representations of temporal information influence evolutionary outcomes and evaluation criteria.

Although the auxiliary human-subjective IEC experiment provided an initial comparison between VLM-based and human-based subjective selection, the number of participants and evaluation terms was limited. Therefore, further studies with a larger and more diverse set of participants, evaluation terms, and VLMs are needed to clarify the robustness, generality, and limitations of using VLMs as fitness functions in evolutionary processes.

By employing the proposed framework, subjective evaluations that were previously difficult to incorporate into evolutionary computation can now be systematically integrated, thereby alleviating bottlenecks caused by the high human labor costs traditionally required for subjective assessment and expanding the scope of feasible applications. This approach could also be integrated with other machine learning approaches for IEC automation, such as LLM-driven genetic operators [17] and auto-encoder-based genotype space construction [18], potentially enabling more flexible and comprehensive automation of evolutionary design systems.

Another important limitation of this study is its dependence on the particular VLM used as the evaluator. In the present experiments, all subjective selections were performed using a 4-bit quantized version of gemma-3-12b-it with a fixed prompt and inference setting. Therefore, the evolved phenotypes should not be interpreted as reflecting universal human subjectivity or general notions of adorableness, weirdness, solemnity, dynamic motion, or motionlessness. Rather, they should be understood as phenotypes favored under the subjective selection pressure induced by this particular VLM in the present experimental setting. The model’s training data, architecture, scale, instruction tuning, quantization, and prompt-dependent interpretation of the evaluation terms may all have influenced the mapping between subjective linguistic expressions and visual or motion features. Accordingly, using a different VLM, a larger model, or a different prompting strategy could lead to different evaluation criteria and, consequently, different evolutionary trajectories.

To examine this model dependency more systematically, future work should compare evolutionary outcomes across multiple VLMs and prompting conditions. For example, it would be valuable to investigate whether the same subjective term leads to similar phenotypes across different models, whether the evaluation criteria extracted from different models overlap, and whether population-level convergence patterns remain stable. Such analyses would make it possible to distinguish phenotypes that are robust across evaluators from those that are specific to a particular model.

Furthermore, the proposed framework has potential applicability to cultural evolutionary computation. In cultural evolution, selection pressure arises from subjective choices made by individuals with diverse perspectives. From this standpoint, rather than fixing the evaluator to a single VLM configuration, a more realistic approach would be to introduce multiple evaluators endowed with different personas, such as occupation, gender, age, or cultural background, and to drive evolution through evaluations derived from these multiple perspectives. Such an extension would not only provide a way to model culturally or socially heterogeneous subjective selection but also make it possible to examine how evolutionary dynamics change when selection pressure is shared, averaged, or conflicted among different evaluators. For instance, phenotypes consistently selected across multiple models or personas could be regarded as relatively robust, whereas phenotypes selected only by a specific model or persona would indicate model- or persona-dependent subjective criteria.

## DATA AVAILABILITY

The code and data supporting the findings of the computational experiments in this study, including videos of representative individuals under different experimental conditions, are publicly available via the Zenodo repository at https://doi.org/10.5281/zenodo.20848566.

## ACKNOWLEDGEMENTS

This study is supported in part by JSPS Topic-Setting Program to Advance Cutting-Edge Humanities and Social Sciences Research JPJS00122674991, JSPS KAKENHI JP24K15103.

**Appendix: Prompt used in the VLM-based evolution experiments**

“{ADVERB}” denotes a subjective evaluation term.

```
# Motion Pattern Evaluation Task

You are an impartial motion analysis specialist who evaluates motion patterns based solely on their intrinsic qualities.
You observe subtle dynamics, rhythm, and temporal flow in object transformations without bias toward spatial positioning.

## Image Description

In this image, there are two frame sequences (top and bottom rows), each containing 4 chronological frames (time steps: t, t+1, t+2, t+3) of tracked red objects from left to right:

- The red object is located in the center of each frame.
- The red object remains completely whole and intact at all times, never splitting or separating into multiple parts, regardless of the passage of time.
- The red object does not rotate.
- Each frame is enclosed by a black border.
- Focus only on how the object's shape changes over time, while maintaining constant size.

## Step-by-Step Analysis Process

### Objective Frame Description

- Top row:
  - Describe the shape in each frame (t, t+1, t+2, t+3).
  - Focus on outline, structure, and distinctive features.

- Bottom row:
  - Describe the shape in each frame (t, t+1, t+2, t+3).
  - Focus on outline, structure, and distinctive features.

### Change Pattern Analysis

- Top row:
  - Describe how the shape transforms between frames.

- Bottom row:
  - Describe how the shape transforms between frames.

### Quality Assessment for "{ADVERB}"

1. Define what "{ADVERB}" means in the context of motion/transformation.
2. Identify relevant characteristics from your observations that relate to this quality.
3. Evaluate which sequence better embodies these characteristics.

### Question

Which sequence moves more {ADVERB} — the top row or the bottom row?
```

```
Support your answer by explaining:
- (1) your definition of "{ADVERB}" in this context
- (2) which observed characteristics are relevant
- (3) how each sequence demonstrates these characteristics
Key Criterion: Provide a one-word term that captures your key criterion for evaluating "{ADVERB}" motion.

At the end of your response, answer with your choice ("top" or "bottom") enclosed in quotation marks.
```

**Prompt used to describe the characteristics of individuals**

```
# Motion Pattern Explanation Task

You are an impartial motion analysis specialist who evaluates motion patterns based solely on their intrinsic qualit
ies.
You observe subtle dynamics, rhythm, and temporal flow in object transformations without bias toward spatial pos
itioning.

## Image Description
In this image, there is a frame sequence (top and bottom rows), each containing 4 chronological frames (time ste
ps: t, t+1, t+2, t+3) of tracked red objects from left to right:

- The red object is located in the center of each frame.
- The red object remains completely whole and intact at all times, never splitting or separating into multiple part
s, regardless of the passage of time.
- The red object does not rotate.
- Each frame is enclosed by a black border.
- Focus only on how the object's shape changes over time, while maintaining constant size.

## Step-by-Step Analysis Process
### Objective Frame Description
- Describe the shape in each frame (t, t+1, t+2, t+3).
- Focus on outline, structure, and distinctive features.

### Change Pattern Analysis
- Describe how the shape transforms between frames.
```